\documentclass[sigconf]{acmart}
\renewcommand\footnotetextcopyrightpermission[1]{} 
\AtBeginDocument{%
  \providecommand\BibTeX{{%
    \normalfont B\kern-0.5em{\scshape i\kern-0.25em b}\kern-0.8em\TeX}}}

\copyrightyear{2023}
\acmYear{2023}
\setcopyright{acmlicensed}\acmConference[FAccT '23]{2023 ACM Conference on Fairness, Accountability, and Transparency}{June 12--15, 2023}{Chicago, IL, USA}
\acmBooktitle{2023 ACM Conference on Fairness, Accountability, and Transparency (FAccT '23), June 12--15, 2023, Chicago, IL, USA}
\acmPrice{15.00}
\acmDOI{10.1145/3593013.3594008}
\acmISBN{979-8-4007-0192-4/23/06}

\definecolor{amethyst}{rgb}{0.6, 0.4, 0.8}

\begin{document}

\title{Domain Adaptive Decision Trees: \\ Implications for Accuracy and Fairness}

\author{Jose M. Alvarez}
\authornotemark[1]
\email{jose.alvarez@sns.it}
\orcid{0000-0001-9412-9013}
\affiliation{%
  \institution{Scuola Normale Superiore, University of Pisa}
  \city{Pisa}
  \country{Italy}
}

\author{Kristen M. Scott}
\authornote{Both authors contributed equally to this research.}
\email{kristen.scott@kuleuven.be}
\orcid{0000-0002-3920-5017}
\affiliation{%
  \institution{KU Leuven, Leuven.AI}
  \city{Leuven}
  \country{Belgium}
}

\author{Bettina Berendt}
\email{berendt@tu-berlin.de}
\orcid{0000-0002-8003-3413}
\affiliation{%
  \institution{TU Berlin, Weizenbaum Institute, KU Leuven}
  \city{Berlin}
  \country{Germany}
}

\author{Salvatore Ruggieri}
\email{salvatore.ruggieri@unipi.it}
\orcid{0000-0002-1917-6087}
\affiliation{%
  \institution{University of Pisa}
  \city{Pisa}
  \country{Italy}
}

\renewcommand{\shortauthors}{}
\renewcommand{\shorttitle}{}

\begin{abstract}
    In uses of pre-trained machine learning models, it is a known issue that the target population in which the model is being deployed may not have been reflected in the source population with which the model was trained. This can result in a biased model when deployed, leading to a reduction in model performance. One risk is that, as the population changes, certain demographic groups will be under-served or otherwise disadvantaged by the model, even as they become more represented in the target population. The field of domain adaptation proposes techniques for a situation where label data for the target population does not exist, but some information about the target distribution does exist. In this paper we contribute to the domain adaptation literature by introducing \textit{domain-adaptive decision trees} (DADT). We focus on decision trees given their growing popularity due to their interpretability and performance relative to other more complex models. With DADT we aim to improve the accuracy of models trained in a source domain (or training data) that differs from the target domain (or test data). We propose an in-processing step that adjusts the information gain split criterion with outside information corresponding to the distribution of the target population. We demonstrate DADT on real data and find that it improves accuracy over a standard decision tree when testing in a shifted target population. We also study the change in fairness under demographic parity and equal opportunity. Results show an improvement in fairness with the use of DADT.
\end{abstract}

\ccsdesc[500]{Computing methodologies~Classification and regression trees}
\ccsdesc[500]{Computing methodologies~Learning under covariate shift}

\keywords{Decision Trees, Information Gain, Domain Adaptation, Covariate Shift, Fairness, folktables}

\received{06 February 2023}
\received[revised]{10 May 2023}
\received[accepted]{}

\maketitle

\section{Introduction}
\label{sec:Introduction}

In uses of pre-trained machine learning models, it is a known issue that the target population in which the model is being deployed may not have been reflected in the data with which the model was trained. There are many reasons why a training set would not match the target population, including sampling bias \cite{suresh2021FrameworkUnderstandingSources}, concept drift \cite{DBLP:journals/kais/GoldenbergW19}, and domain shift \cite{DBLP:journals/kais/GoldenbergW19}. This situation can lead to a reduction in model performance in the target domain. One risk is that, as the demographic distribution of the population changes, certain groups will be under-served by model performance, even as they become more represented in the target population: a type of representation bias \cite{suresh2021FrameworkUnderstandingSources}. Lack of representation, or invisibility, of this kind can be unfair, and adequate visibility can be a prerequisite for fairness \cite[Chapter~4]{dignazio2020DataFeminism}. A classic example is that of female and darker-skinned people being underrepresented in computer-vision datasets, hence scarcely visible to the learning algorithm, with consequences like high error rates in facial recognition and consequent denials of benefits (such as authentication) or imposition of harms (such as arrests) \cite{DBLP:conf/fat/BuolamwiniG18}. One, often advisable, approach for dealing with this is to train a new model with updated or improved training data. However, in the case of supervised learning, this may not be possible, as label information for these additional members of the target population may not yet exist. Additionally, while collection of representative data is very important, it does come at a cost, including a time cost, so that some shift in the target is likely to occur before updated data is collected or a shift is even identified. The field of domain adaptation proposes techniques for addressing these situations \cite{Redko2020_DASurvey}. 

In this paper we contribute to the domain adaptation literature by introducing \textit{domain-adaptive decision trees} (DADT). With DADT we aim to improve accuracy of decision tree models trained in a source domain (or training data) that differs from the target domain (or test data), as it may occur when we do not have labeled instances for the target domain. We do this by proposing an in-processing step that adjusts the information gain (IG) split criterion with outside information in the form of unlabeled data, corresponding to the distribution of the target population we aim for. The approach works by adapting probability estimation to the target domain and, thus, making parts of the feature space more visible to the learning algorithm. We investigate the conditions in which this strategy can lead to increases in performance, fairness, or both.

As an illustrative example, consider the case of a sports retail store looking to target new clients ($D_T$) using what it knows about its current clients ($D_S$) based on geographical regions. The store only has information on the purchasing habits ($Y$) of $D_S$ . Imagine that the store wants to use a classifier to inform its inventory on women's football shoes. If the two client populations differ by region, which is likely, the classifier trained on $D_S$ and intended to predict purchasing patterns ($\hat{Y}$) on $D_T$ could lead to biased predictions when used. For instance, if there is less demand for women’s football shoes in the source region relative to the target region, the classifier could underestimate the stocks of women's football shoes needed, under-serving the potential new clients. This could lead to lower service or higher prices for some social groups, and the lost opportunity by the store to gain or even retain customers in the target region. To break such feedback loops, the store could improve the classifier by amplifying some of the knowledge about football shoes purchases in the source region. It could, for instance, use knowledge about the demographics in the target region to better approximate the demand for football shoes by women. 

We focus on decision trees for domain shift because decision trees are accessible, interpretable, and well-performing classification models that are commonly used. In particular, we study decision trees rather than more complex classifiers when using tabular data for three reasons. First, these models are widely available across programming languages and are standard in industry and academic communities \cite{Rudin2016_KeyNotePapis}. Second, these models are inherently transparent \cite{Rudin2019_StopExplainingML}, which may facilitate the inclusion of stakeholders in understanding and assessing model behaviour. Third, ensembles of these models still outperform the deep learning models on tabular data~\cite{DBLP:conf/nips/GrinsztajnOV22}. For these reasons, and as proposed AI regulations include calls for explainable model behaviour \cite{EU_AIAct2, USA_AIBill}, decision trees are a relevant choice when training a classifier and it is therefore important to address issues specific to them.

There are different types of domain shift \cite{Quinonero2009} and they have different implications for suitable interventions. We focus on the \textit{covariate shift} case of domain shift. This is the case where only the distribution of the attributes change between the source and target, not the relationship between the attributes and the label. 

In Section~\ref{sec:ProblemSetting}, we introduce the problem setting as a domain adaptation problem, focusing on the covariate shift type of this problem. We also present the necessary background before presenting our proposed intervention to the information gain and introduce the domain adaptive decision trees in Section \ref{sec:DADT}.

Then in Section \ref{sec:Experiments}, we present the results of our experiments. In the experiments reported we utilize the \textit{ACSPublicCoverage} dataset---an excerpt of US Census data \cite{dingRetiringAdultNew2021}, with the prediction task of whether or not a low income individual is covered by public health coverage. The dataset provides the same feature sets for each of the US states. This design allows us to set up an experimental scenario that mirrors our retail example of having no labeled data for the target domain, but some knowledge of the distribution of the attributes in the target domain.

With these experiments we aim not only to improve overall accuracy, but also to produce sufficient accuracy for different demographic groups. This is important because the distribution of these groups may be different in the target population and even shift over time in that population. For example, Ding et al. \cite{dingRetiringAdultNew2021} found that na\"{\i}vely implementing a model trained on data from one US state and using it in each of the other states resulted in unpredictable performance in both overall accuracy and in \emph{demographic parity} (a statistically defined metric of model fairness performance based on treatment of members of a selected demographic group compared to members of another demographic group). We therefore also test the impact of our intervention on the results of a post-processing fairness intervention \cite{DBLP:conf/nips/HardtPNS16}, which we measure using two common fairness metrics: \textit{demographic parity} and \textit{equal opportunity} \cite[Chapter~3]{barocas-hardt-narayanan2019}.

We examine those results in relation to the covariate shift assumption between source and target populations. We see that our intervention leads to an increase in accuracy when the covariate shift assumption holds. The related work in Section~\ref{sec:RelatedWork} situates our approach in the literature on domain adaptation in decision trees and adjusting the information gain of decision trees. Section \ref{sec:Closing} closes and gives an outlook on future work.

{\bf Research Ethics:} We use data from an existing benchmark designed for public use and work with aggregated or anonymized data only, thus complying with applicable legal and  ethical rules, and we disclose all details of our method in line with the transparency mandates of the ACM Code of Ethics. 

%
%

%
\section{Problem Setting}
\label{sec:ProblemSetting}

Let $\mathbf{X}$ denote the set of discrete/continuous \textit{predictive attributes}, $Y$ the \textit{class attribute}, and $f$ the \textit{decision tree classifier} such that $\hat{Y}=f(\mathbf{X})$ with $\hat{Y}$ denoting the \textit{predicted class attribute}. We assume a scenario where the population used for training $f$ (the \textit{source domain} $D_S$) is not representative of the population intended for $f$ (the \textit{target domain} $D_T$). Formally, we write it as $P_S(\mathbf{X}, Y) \neq P_T(\mathbf{X}, Y)$, where $P_S(\mathbf{X}, Y)$ and $P_T(\mathbf{X}, Y)$, respectively, denote the source and target domain joint probability distributions. We tackle this scenario as a \textit{domain adaptation} (DA) problem \cite{Redko2020_DASurvey} as it allows us to formalize the difference between distributions in terms of distribution shifts. There are three types of distribution shifts in DA: covariate, prior probability, and dataset shift. Here, we focus on \textit{covariate shift} \cite{Quinonero2009, DBLP:journals/pr/Moreno-TorresRACH12, DBLP:conf/icml/ZhangSMW13} in which the conditional distribution of the class, $P(Y|\mathbf{X})$, remains constant but the marginal distribution of the attributes, $P(\mathbf{X})$, changes across the two domains:
\begin{equation}
\label{eq:CovariateShiftDefinition}
    P_S(Y | \mathbf{X}) = P_T(Y | \mathbf{X}) \; \; \text{but} \; \;
    P_S(\mathbf{X}) \neq P_T(\mathbf{X})
\end{equation}
We focus on covariate shift because we assume, realistically, to have some access only to the predictive attributes $\mathbf{X}$ of the target domain.\footnote{The other two settings require information on $Y$ being available in $D_T$, with \textit{prior probability shift} referring to cases where the marginal distribution of the class attribute changes, $P_S(\mathbf{X}|Y) = P_T(\mathbf{X}|Y)$ but $P_S(Y) \neq P_T(Y)$, and \textit{dataset shift} referring to cases where neither covariate nor prior probability shifts apply but the joint distributions still differ, $P_S(\mathbf{X}, Y) \neq P_T(\mathbf{X}, Y)$.} Under this \textit{unsupervised setting}, we picture a scenario where a practitioner needs to train $f$ on $D_S$ to be deployed on $D_T$. Aware of the potential covariate shift, the practitioner wants to avoid training a biased model relative to the target domain that could result in poor performance on $\hat{Y}$.

What can be done here to address the DA problem depends on what is known about $P_T(\mathbf{X})$. In the ideal case in which we know the whole covariate distribution $P_T(\mathbf{X})$, being under \eqref{eq:CovariateShiftDefinition} allows for computing the full joint distribution due to the multiplication rule of probabilities:
\begin{equation}
\label{eq:chain}
   P_T(Y, \mathbf{X}) = P_T(Y | \mathbf{X}) \cdot P_T(\mathbf{X}) = P_S(Y | \mathbf{X}) \cdot P_T(\mathbf{X})
\end{equation}
where we can exchange $P_T(Y|\mathbf{X})$ for $P_S(Y|\mathbf{X})$, which is convenient as we know both $Y$ and $\mathbf{X}$ in $D_S$. In reality, however, the right-hand-side of \eqref{eq:chain} can be known to some extent due to three issues:
\begin{enumerate}
    \item[(P1)] $P_T(\mathbf{X})$ is not fully available, meaning the marginal distributions of some of the attributes $X \in \mathbf{X}$ are known;
    \item[(P2)] $P_T(Y | \mathbf{X}) \approx P_S(Y | \mathbf{X})$ but not equal, meaning the covariate shift holds in a relaxed form; and
    \item[(P3)] $P_T(\mathbf{X})$ and $P_S(Y | \mathbf{X})$ are estimated given sample data from the respective populations, and, as such, the estimation can have some variability. 
\end{enumerate}
Issue P3 is pervasive in statistical inference and machine learning. We do not explicitly\footnote{We tackle it implicitly through the Law of Large Numbers by restricting to estimation of probabilities in contexts with a minimum number of instances. This is managed in decision tree learning by a parameter that stops splitting a node if the number of instances at a node is below a minimum threshold.} consider it in our problem statement. Therefore, the main research question that we intend to address in this paper is:

\smallskip
\textit{RQ1. With reference to the decision tree classifier, which type and amount of target domain knowledge (issue P1) help reduce the loss in accuracy at the variation of relaxations of covariate shift (issue P2)?}

\smallskip
As domain shift can have a detrimental impact on performance of the model for some demographic groups over others, a subsequent question to address in this paper is:

\smallskip
\textit{RQ2. How does the loss in accuracy by the decision tree classifier, based on the issues P1 and P2, affect a fairness metric used for protected groups of interest in the target domain?}

\smallskip
The knowledge relevant for \eqref{eq:chain} and RQ1 and RQ2 is bounded by two border cases.
\textbf{\textit{No target domain knowledge:}} it consists of training $f$ on the source data and using it on the target data without any change or correction. Formally, we estimate $P_T(\mathbf{X})$ as $P_S(\mathbf{X})$ and $P_T(Y | \mathbf{X})$ as $P_S(Y | \mathbf{X})$.
\textbf{\textit{Full target domain knowledge:}} it consists of training $f$ on the source data and using it on the target data, but exploiting full knowledge of $P_T(\mathbf{X})$ in the learning algorithm to replace $P_S(\mathbf{X})$.
\textbf{\textit{Partial target domain knowledge:}} consequently, the in-between case consists of training a decision tree on the source data and using it on the target data, but exploiting partial knowledge of $P_T(\mathbf{X})$ in the learning algorithm and complementing it with knowledge of $P_S(\mathbf{X})$.

The form of partial knowledge depends on the information available on $\mathbf{X}$, or subsets of it. Here, we consider a scenario where for $\mathbf{X}' \subseteq \mathbf{X}$, an estimate of $P(\mathbf{X}')$ is known  only for $|\mathbf{X}'| \leq 2$ (or $|\mathbf{X}'| \leq 3$), namely we assume to know bi-variate (resp., tri-variate) distributions only, but not the full joint distribution. This scenario occurs, for example, when using cross-tabulation data from official statistics. We specify how to exploit the knowledge of $P_T(\mathbf{X})$ for a decision tree classifier in Section~\ref{sec:DADT}, introducing what we refer to as a \textit{domain-adaptive decision tree} (DADT). We introduce the required technical background in the remainder of this section. 

\subsection{Decision Tree Learning}
\label{sec:ProblemSetting.DecisionTreeLearning}
 
Top-down induction algorithms grow a decision tree classifier \cite{Hastie2009_ElementsSL} from the root to the leaves. At each node, either the growth stops producing a leaf, or a split condition determines child nodes that are recursively grown. Common stopping criteria include node purity (all instances have the same class value), data size (the number of instances is lower than a threshold), and tree depth (below a maximum depth allowed). Split conditions are evaluated based on a split criterion, which selects one of them or possibly none (in this case the node becomes a leaf).

We assume binary splits of the form:\footnote{There are other forms of binary splits, as well as multi-way and multi-attribute split conditions \cite{Kumar2006}.} $X=t$ for the left child and $X\neq t$ for the right child, when $X$ is a discrete attribute; or $X\leq t$ for the left child and $X > t$ for the right child, when $X$ is a continuous attribute. We call $X$ the \textit{splitting attribute}, and $t \in X$ the \textit{threshold value}. Together they form the \textit{split condition}. Instances of the training set are passed from a node to its children by partitioning them based on the split condition. The conjunction of split conditions from the root to the current node being grown is called the \textit{current path} $\varphi$. It determines the instances of the training dataset being considered at the current node. The predicted probability of class $y$ at a leaf node is an estimation of $P(Y=y|\varphi)$ obtained by the relative frequency of $y$ among the instances of the training set reaching the leaf or, equivalently, satisfying $\varphi$.

\subsection{The Information Gain Split Criterion}
\label{sec:ProblemSetting.Entropy}

We focus on the information gain split criterion. It is, along with Gini, one of the standard split criteria used. It is also based on information theory via entropy \cite{Cover1999ElementsIT}, which links the distribution of a random variable to its information content. The \textit{entropy} ($H$) measures the information contained within a random variable based on the uncertainty of its events. The standard is \textit{Shannon's entropy} \cite{DBLP:conf/icaisc/MaszczykD08} where we define $H$ for the class random variable $Y$ at $\varphi$ as:
\begin{equation}
\label{eq:ShannonEntropy}
        H(Y|\varphi) = \sum_{y \in Y} - P(Y=y|\varphi) \log_2(P(Y=y|\varphi))
\end{equation}
where $-\log_2(P(Y=y|\varphi))=I(y|\varphi)$ represents the \textit{information} ($I$) of $Y=y$ at current path $\varphi$. Therefore, entropy is the expected information of the class distribution at the current path. Intuitively, the information of class value $y$ is inversely proportional to its probability $P(Y=y|\varphi)$. The more certain $y$ is, reflected by a higher $P(Y=y|\varphi)$, the lower its information as $I(y|\varphi)$ (along with its contribution to $H(Y|\varphi)$). The general idea is that there is little new information to be learned from an event that is certain to occur. 

The \textit{information gain} ($IG$) for a split condition is the difference between the entropy at a node and the weighted entropy at the child nodes determined by the split condition $X$ and $t$ under consideration. $IG$ uses \eqref{eq:ShannonEntropy} to measure how much information is contained under the current path. For a discrete splitting attribute $X$ and threshold $t$, we have:
\begin{multline}
\label{eq:BasicIGd}
    IG(X, t | \varphi) =  H(Y|\varphi) - P(X=t|\varphi) H(Y|\varphi, X=t) \\ - P(X\neq t|\varphi) H(Y|\varphi, X \neq t)
\end{multline}
and for a continuous splitting attribute $X$ and threshold $t$:
\begin{multline}
\label{eq:BasicIGc}
    IG(X, t | \varphi) =  H(Y|\varphi) - P(X\leq t|\varphi) H(Y|\varphi, X\leq t) \\ - P(X> t|\varphi) H(Y|\varphi, X > t)
\end{multline}
where the last two terms in each \eqref{eq:BasicIGd} and \eqref{eq:BasicIGc} represent the total entropy obtained from adding the split condition on $X$ and $t$ to $\varphi$.\footnote{Formally, together these last two terms represent the conditional entropy $H(Y,X|\varphi)$ written for the binary split case we are considering such that:
\begin{equation*}
\label{eq:CondShannonEntropy}
        H(Y, X|\varphi) = 
        \sum_{x \in X} - P(X=x|\varphi) \sum_{y \in Y} P(Y=y|\varphi, X=x) \log_2(P(Y=y|\varphi, X=x))
\end{equation*}
}
The selected split attribute and threshold are those with maximum $IG$, namely $\mathit{arg\,max}_{X, t}\,IG(X, t | \varphi)$.

\subsection{On Estimating Probabilities}
\label{sec:ProblemSetting.EstProb}

Probabilities and, thus, $H$~\eqref{eq:ShannonEntropy} and $IG$~\eqref{eq:BasicIGd}--\eqref{eq:BasicIGc} are defined for random variables. As the decision tree grows, the probabilities are estimated on the subset of the training set $D$ reaching the current node satisfying $\varphi$ by frequency counting:
\begin{multline}
\label{eq:EstProb}
        \hat{P}(X=t \, | \, \varphi) = \frac{|\{w \in D\ |\ \varphi(w) \wedge w[X] = t\}| }{|\{w \in D\ |\ \varphi(w) \}|} 
    \, , \, \\
     \hat{P}(Y=y \, | \, \varphi) =  \frac{|\{w \in D\ |\ \varphi(w) \wedge w[Y] = y\}| }{|\{w \in D\ |\ \varphi(w) \}|}
\end{multline}
The denominator represents the number of instances $w$ in $D$ that satisfy the condition $\varphi$ (written $\varphi(w)$) and the numerator the number of those instances that further satisfy $X=t$ (respectively, $Y=y$). We use the estimated probabilities \eqref{eq:EstProb} to estimate $H$ and $IG$.

The hat in \eqref{eq:EstProb} differentiates the estimated probability, $\hat{P}$, from the population probability, $P$. Frequency counting is supported by the Law of Large Numbers. Assuming that the training set $D$ is an \textit{i.i.d.} sample from the $P$ probability distribution, we expect for $\hat{P}(X=t|\varphi) \approx P(X=t \, | \, \varphi)$ and $\hat{P}(Y=y|\varphi) \approx P(Y=y \, | \, \varphi)$ as long as we have enough training observations in $D$, which is often the case when training $f$. A key issue is whether $D$ is representative of the population of interest. This is important as $\hat{P}$ will approximate the $P$ behind $D$.

When training any classifier, the key assumption is that the $P$ probability distribution is the same for the data used for growing the decision tree (the training dataset) and for the data on which the decision tree makes predictions (the test dataset). Under covariate shift \eqref{eq:CovariateShiftDefinition} this assumption does not hold. Instead, the training dataset belongs to the source domain $D_S$ with  probability distribution $P_S$ and the test dataset belongs to the target domain $D_T$ with probability distribution $P_T$, such that $P_T \neq P_S$. To stress this point, we use the name \textit{source data} for training data sampled from $D_S$ and \textit{target data} for training data sampled from $D_T$. The estimated probabilities \eqref{eq:EstProb}, and the subsequent estimations for $H$ \eqref{eq:ShannonEntropy} and $IG$ \eqref{eq:BasicIGd}--\eqref{eq:BasicIGc} based on source data alone can be biased, in statistical terms, relative to the intended target domain (recall issue P1). This can result, among other issues, in poor model performance from the classifier. This is why we propose extending \eqref{eq:EstProb} by embedding target-domain knowledge into these estimated probabilities.

Measuring the distance between the probability distributions $P_S$ and $P_T$ is relevant for detecting distribution shifts. We resort to the \textit{Wasserstein distance} $W$ between two probability distributions to quantify the amount of covariate shift and the robustness of target domain knowledge. See
Appendix~A.1
for details. Under covariate shift \eqref{eq:CovariateShiftDefinition}, it is assumed that $P_S(Y|\mathbf{X}) = P_T(Y|\mathbf{X})$, which allows one to focus on the issue of $P_S(\mathbf{X}) \neq P_T(\mathbf{X})$. This equality is often not verified in practice. We plan to use $W$, along with an approximation of $P_T(Y|\mathbf{X})$ (since $Y$ is unavailable in $D_T$), to measure the distance between these two conditional probabilities to ensure that our proposed embedding with target domain knowledge is impactful (issue P2). Measuring this distance will allow us to evaluate how relaxations of $P_S(Y|\mathbf{X}) = P_T(Y|\mathbf{X})$ affect the impact of our proposed target domain embedding.

%
%

%
\section{Domain-Adaptive Decision Trees}
\label{sec:DADT}

We present our approach for addressing covariate shift by embedding target domain knowledge when learning the decision tree classifier. We propose an \textit{in-processing step} under the information gain split criterion, motivating what we refer to as \textit{domain-adaptive decision trees} (DADT) learning. As discussed in Section~\ref{sec:ProblemSetting}, when growing the decision tree, the estimated probabilities \eqref{eq:EstProb} used for calculating $H$ \eqref{eq:ShannonEntropy} and thus $IG$ \eqref{eq:BasicIGd}--\eqref{eq:BasicIGc} at the current path $\varphi$ are derived over a training dataset, which is normally a dataset over the source domain $D_S$. For the split condition $X=t$, it follows that $\hat{P}(X=t | \varphi) \approx P_S(X=t | \varphi)$, which is an issue under covariate shift. We instead want that $\hat{P}(X=t | \varphi) \approx P_T(X=t | \varphi)$. We propose to embed in the learning process knowledge from the target domain $D_T$, reducing the potential bias in the estimation of the probabilities and, in turn, reducing the bias of the trained decision classifier.

\subsection{Embedding Target Domain Knowledge}
\label{sec:DADT:embedding}

There are two probability forms that are to be considered when growing a decision tree for the current path $\varphi$: $P(X=t|\varphi)$  in \eqref{eq:BasicIGd} (and, respectively, $P(X\leq t|\varphi)$ in \eqref{eq:BasicIGc}) and $P(Y=y|\varphi)$ in \eqref{eq:ShannonEntropy}--\eqref{eq:CondShannonEntropy}. In fact, the formulas of entropy and information gain only rely on those two probability forms, and on the trivial relation $P(X\neq t|\varphi) = 1 - P(X=t|\varphi)$ for discrete attributes (and, respectively, $P(X> t|\varphi) = 1 - P(X\leq t|\varphi)$ for continuous attributes). It follows that we can easily estimate $\hat{P}_S(X=t|\varphi)$ and $\hat{P}_S(Y=y|\varphi)$ using the available source domain knowledge.

\subsubsection{Estimating $P(X=t|\varphi)$} 
We assume that some target domain knowledge is available, from which we can estimate $\hat{P}_T(X|\varphi)$, in the following cases:\footnote{
Actually, since we consider $x$ in the (finite) domain of $X$ the two forms are equivalent, due to basic identities $P_T(X\leq x|\varphi) = \sum_{X \leq x} P_T(X = x|\varphi)$ and $P_T(X = x|\varphi) = P_T(X\leq x|\varphi) - P_T(X\leq x'|\varphi)$, where $x'$ is the element preceding $x$ in the domain of $X$. Moreover, by definition of conditional probability, we have $P(X=t|\varphi) = P(X=t, \varphi)/P(\varphi)$ and then target domain knowledge boils down to estimates of probabilities of conjunction of equality conditions. Such form of knowledge is, for example, provided by cross-tables in official statistics data.
}
\begin{multline}
\label{eq:tdk} 
    \hat{P}_T(X=x|\varphi) \approx P_T(X=x|\varphi) \mbox{\rm\ for $X$ discrete,} \\ \hat{P}_T(X\leq x|\varphi) \approx P_T(X\leq x|\varphi) \mbox{\rm\ for $X$ continuous} 
\end{multline}
In case $\hat{P}_T(X=x|\varphi)$ is not directly available in the target domain knowledge $D_T$, we adopt an affine combination for discrete and continuous attributes using the source domain knowledge $D_S$:
\begin{eqnarray}
\label{eq:aff}
    & \hat{P}(X=x|\varphi) = \alpha \cdot  \hat{P}_S(X=x|\varphi)  + (1-\alpha) \cdot \hat{P}_T(X=x|\varphi') \\
\label{eq:aff2}
    & \hat{P}(X\leq x|\varphi) = \alpha \cdot  \hat{P}_S(X\leq x|\varphi)  + (1-\alpha) \cdot \hat{P}_T(X\leq x|\varphi')
\end{eqnarray}
where $\varphi'$ is a maximal subset of split conditions in $\varphi$ for which $\hat{P}_T(X=x|\varphi')$ is in the target domain knowledge, and $\alpha \in [0, 1]$ is a tuning parameter to be set. In particular, setting $\alpha=1$ boils down to estimating probabilities based on the source data only. With such assumptions, $P(X=t|\varphi)$ in \eqref{eq:BasicIGd} (respectively $P(X\leq t|\varphi)$ in \eqref{eq:BasicIGc}) can be estimated as $\hat{P}(X=t|\varphi)$ (resp. 
$\hat{P}(X \leq t|\varphi)$) to derive $IG$. 

\subsubsection{Estimating $P(Y=y|\varphi)$} 
Let us consider the estimation of $P(Y=y|\varphi)$ in \eqref{eq:ShannonEntropy} over the target domain. Since $Y$ is unavailable in $D_T$, it is legitimate to ask whether $P(Y=y|\varphi)$ is the same probability in the target as in the source domain when growing the decision tree classifier. If yes, then we would simply estimate $P_T(Y=y|\varphi) \approx \hat{P}_S(Y=y|\varphi)$. Unfortunately, the answer is no. Recall that the covariate shift assumption \eqref{eq:CovariateShiftDefinition} states that $P_S(Y|\mathbf{X}) = P_T(Y|\mathbf{X})$, namely that the probability of $Y$ conditional on fixing \textit{all of the} variables in $\mathbf{X}$ is the same in the source and target domains:
\begin{align}
\label{eq:covshift}
    \forall \mathbf{x} \in \mathbf{X}, \forall y \in Y, \ P_S(Y=y|\mathbf{X=x}) = P_T(Y=y|\mathbf{X=x})
\end{align}
However, this equality may not hold when growing the tree because the current path $\varphi$ does not necessarily fix all of the $\mathbf{X}$'s, i.e., \eqref{eq:covshift} does not necessarily imply $\forall \varphi \; P_T(Y=y|\varphi) = P_S(Y=y|\varphi)$. This situation, in fact, is an instance of  Simpson's paradox \cite{Simpson1951_Interpretation}. We show this point with Example~A.1 in Appendix~A.2. 
We rewrite $P_T(Y=y|\varphi)$ using the law of total probability as follows:
\begin{multline}
\label{eq:rewry}
    P_T(Y=y|\varphi) 
        = \sum_{\mathbf{x} \in \mathbf{X}} P_T(Y=y|\mathbf{X}=\mathbf{x}, \varphi) \cdot P_T(\mathbf{X}=\mathbf{x}|\varphi) \\
        = \sum_{\mathbf{x} \in \mathbf{X}} P_S(Y=y|\mathbf{X}=\mathbf{x}, \varphi) \cdot P_T(\mathbf{X}=\mathbf{x}|\varphi) 
\end{multline}
where the final equation exploits the covariate shift assumption \eqref{eq:covshift} when it holds for a current path $\varphi$. Now, instead of taking $P_S(Y|\varphi)=P_T(Y|\varphi)$ for granted, which we should not do under DADT learning, we rewrite $P_T(Y=y|\varphi)$ in terms of probabilities over source domain, $P_S(Y=y|\mathbf{X}=\mathbf{x}, \varphi)$, and target domain, $P_T(\mathbf{X}=\mathbf{x}|\varphi)$, knowledge. 

Varying $\mathbf{x} \in \mathbf{X}$ over all possible combination as stipulated in \eqref{eq:rewry}, however, is not feasible in practice as it would require extensive target domain knowledge to estimate $P_T(\mathbf{X}=\mathbf{x}|\varphi)$ $\forall \mathbf{x} \in \mathbf{X}$. This would still be a practical issue in the ideal case in which we have a full sample of the target domain, as it would require the sample to be large enough for observing each value $\mathbf{x} \in \mathbf{X}$ in $D_T$. We approximate \eqref{eq:rewry} by varying values with respect to a \textit{single attribute} $X_w \in \mathbf{X}$ and relying on \eqref{eq:aff} for an estimate of $P_T(X_w=x|\varphi)$. Let us then define the estimate of $P(Y=y|\varphi)$ as:
\begin{align}
\label{eq:wpyphi}
    \hat{P}(Y=y|\varphi) = \sum_{x \in X_w} \hat{P}_S(Y=y|X_w=x, \varphi) \cdot \hat{P}(X_w=x|\varphi) 
\end{align}
where we now use target domain knowledge only about $X_w$ instead of spanning the entire attribute space $\mathbf{X}$. 

To account for the above instance of Simpson's paradox, the attribute $X_w$ should be chosen such that  $\hat{P}_S(Y=y|X_w=x,\varphi) \approx P_T(Y=y|X_w=x,\varphi)$. Such an attribute $X_w$, however, may be specific to the current path $\varphi$. Hence, we only consider the empty $\varphi$, and  choose $X_w$ such that the average distance between $\hat{P}_S(Y|X_w=x)$ and an estimate $\hat{P}_T(Y|X_w=x)$ of $P_T(Y|X_w=x)$  is minimal:
\begin{multline}
\label{eq:wpy}
    X_w = \arg \min_{X} \mathcal{W}(X) \ \mbox{\rm where\ }\\  
    \mathcal{W}(X) = \sum_{x \in X} W(\hat{P}_S(Y|X=x),\hat{P}_T(Y|X=x)) \cdot \hat{P}_T(X=x)
\end{multline}
$\mathcal{W}(X)$ is the average Wasserstein distance between $\hat{P}_S(Y|X)$ and $\hat{P}_T(Y|X)$. In terms of target domain knowledge, computing \eqref{eq:wpy} requires knowledge of $\hat{P}_T(Y|X)$, an estimate
of the conditional distribution of the class in the target domain. 

In \eqref{eq:wpy}, we depart slightly from our assumption that no knowledge is available of $Y$ in $D_T$. If calculation of  \eqref{eq:wpy} is not feasible, we assume some expert input on an attribute $X$ such that $\hat{P}_S(Y=y|X=x) \approx \hat{P}_T(Y=y|X=x)$, as a way to minimize the first term of the summation (\ref{eq:wpy}). In such a case, we do not actually compute $\mathcal{W}(X)$. 

To summarize, we use $X_w$ as from \eqref{eq:wpy} (or as provided by a domain expert) to derive \eqref{eq:wpyphi} as an empirical approximation to \eqref{eq:rewry}. This is how we estimate $P(Y|\varphi)$ over the target domain.

\subsection{How Much Target Domain Knowledge?}
\label{sec:DADT:knowledge}

We can now formalize the range of cases based on the availability of $P_T(\mathbf{X})$ described in Section \ref{sec:ProblemSetting}. Under \textit{\textbf{no target domain knowledge}}, we have no information available on $D_T$, which means that $\hat{P}(X=x|\varphi) = \hat{P}_S(X=x|\varphi)$. This amounts to setting  $\alpha=1$ in \eqref{eq:aff} and \eqref{eq:aff2}, and, whatever $X_w$ is, \eqref{eq:wpyphi} boils down to $\hat{P}(Y=y|\varphi) = \hat{P}_S(Y=y|\varphi)$. In short, both probability estimations boil down to growing the DADT classifier using the source data $D_S$ without any modification or, simply, growing a standard decision tree classifier. Similarly, under \textit{\textbf{full target domain knowledge}} we have target domain knowledge for all attributes in $D_T$ along with enough instances to estimate both probabilities. This amounts to setting $\alpha=0$ in \eqref{eq:aff}--\eqref{eq:aff2}, and to know which attribute $X_w$ minimizes \eqref{eq:wpy}. 

The full target domain knowledge is the strongest possible assumption within our DADT approach, but not in general. For validation purposes (Section~\ref{sec:Experiments}), we move away from our unsupervised setting and assume $Y \in D_T$ to set up an \textit{additional baseline} under the full knowledge of $D_T$: \textit{\textbf{target-to-target baseline.}} In this scenario, the decision tree is grown \textit{and} tested exclusively on the target data. Such a scenario does not require covariate shift, since probabilities $P(X=t|\varphi)$ and $P(Y=y|\varphi)$ are estimated directly over the target domain. This is \textit{the ideal case} as we train the classifier on the intended population.

Finally, under \textit{\textbf{partial target domain knowledge}} we consider cases where we have access to estimates of $P(\mathbf{X}')$ only for some subsets $\mathbf{X}' \subseteq \mathbf{X}$. This allows us to estimate $P(X=x|\varphi)$ only if $X$ and the variables in $\varphi$ are in one of those subsets $\mathbf{X}'$. When the target domain information is insufficient, DADT resorts to the source domain information in \eqref{eq:aff}--\eqref{eq:aff2} by an affine combination of both. The weight $\alpha$ in such an affine combination should be set proportional to the contribution of the source domain information. We refer to Section \ref{sec:AccuracyResults} for our experimental setting of $\alpha$.

%
%

%
\section{Experiments: State Public Coverage}
\label{sec:Experiments}

We consider the \textit{ACSPublicCoverage} dataset---an excerpt from the 2017 US Census data \cite{dingRetiringAdultNew2021}---that provides the same feature sets for different geographical regions based on the US states, which may have different distributions. This allows us to examine the impact of our method given a wide range of distribution shifts. We utilize the prediction task, constructed by the dataset creators, of whether or not a low-income individual is covered by public health coverage. 

Inspired by this experimental setting, we imagine a task where a public administrator wants to identify individuals who do not receive the public benefits they are entitled to. Information about who does and does not receive these benefits, however, is only available for a population different from the target population: for example, the population from another state. This administrator is likely to have some information about the target population distribution; information that they realistically may have on population breakdown by demographics such as age, race and gender. To address RQ1 we now test whether, with DADT, we can utilize that information to train an improved model in the new state, compared to blindly applying a model trained in the other state. Additionally, we address RQ2 by testing the impact of using DADT instead of standard decision trees on two fairness metrics: demographic parity and equal opportunity.

\subsection{Experimental Setup}
\label{sect:expsetup}

The design of the \textit{ACSPublicCoverage} dataset allows us to set up a scenario that mirrors our example of the retail store in Section~\ref{sec:Introduction}: we have unlabeled data for the target domain, but some knowledge of the (unconditional) distribution of the target domain.
%
Here, however, we extend the scenario by having access to the labeled data for each state. We note that the implementation of the DADT does not require this information, but
%
we utilize our access to the target domain labeled data in Section~\ref{sec:Experiments.CovariateShift} to test our assumption that DADTs are suitable for addressing covariate shift. 
Given the dataset design, we are able to utilize the distribution of the predictive attributes in the target domain as our source of outside knowledge, to adjust the information gain calculation. Unless otherwise stated, we consider the attributes: SCHL (educational attainment), MAR (marital status), AGEP (age), SEX (male or female), CIT (citizenship status), RAC1P (race), with AGEP being continuous and all others discrete. Data was accessed through the Python package Folktables.\footnote{\url{https://github.com/zykls/folktables}}

We consider pairs of source and target datasets consisting of data from different US states, with a model trained in each of the fifty states being tested on every state, for a total of 2500 train / test pairs. The decision trees are all trained on 75\% of source data $D_S$, and tested on 25\% of the target data $D_T$. Stopping criteria include the following: a node must have at least 5\% of the training data, and not all instances have the same class value (purity level set to 100\%), the maximum tree depth is 8.\footnote{The code, data, and run are available at \url{https://github.com/nobias-project/domain-adaptive-trees}.}

To address RQ2, in particular, we undertake a post-processing approach to fairness based on the known link between a model's performance and its fairness \cite{DBLP:conf/icml/DuttaWYC0V20, DBLP:journals/ijis/ValdiviaSC21, DBLP:conf/sigecom/LiangLM22}. The public administrator wants to evaluate the performance of the trained classifier on certain demographic groups in the target population. The administrator thus resorts to applying a post-processing method around the classifier that adjusts the predictions under the chosen fairness metric. In practice, this comes down to using a wrapper function based on \cite{DBLP:conf/nips/HardtPNS16}. 

We focus on this model agnostic post-processing fairness intervention to measure the impact of DADT on a fairness intervention. Post-processing methods rely on the non-DA setting, meaning that $P_S(\mathbf{X}, Y) = P_T(\mathbf{X}, Y)$. Classifiers are a statement on the joint probability distribution of the training data. Under DA, post-processing methods are essentially only modifying $P_S(\mathbf{X}, Y)$. Granted the user trains an oracle-like standard decision tree, the issue remains that the post-processing fairness intervention would only be addressing issues on the source and not the target population. Therefore, DADT is expected to positively affect the fairness measure.

\begin{figure*}[t]
  \centering
  \includegraphics[width=0.41\textwidth]{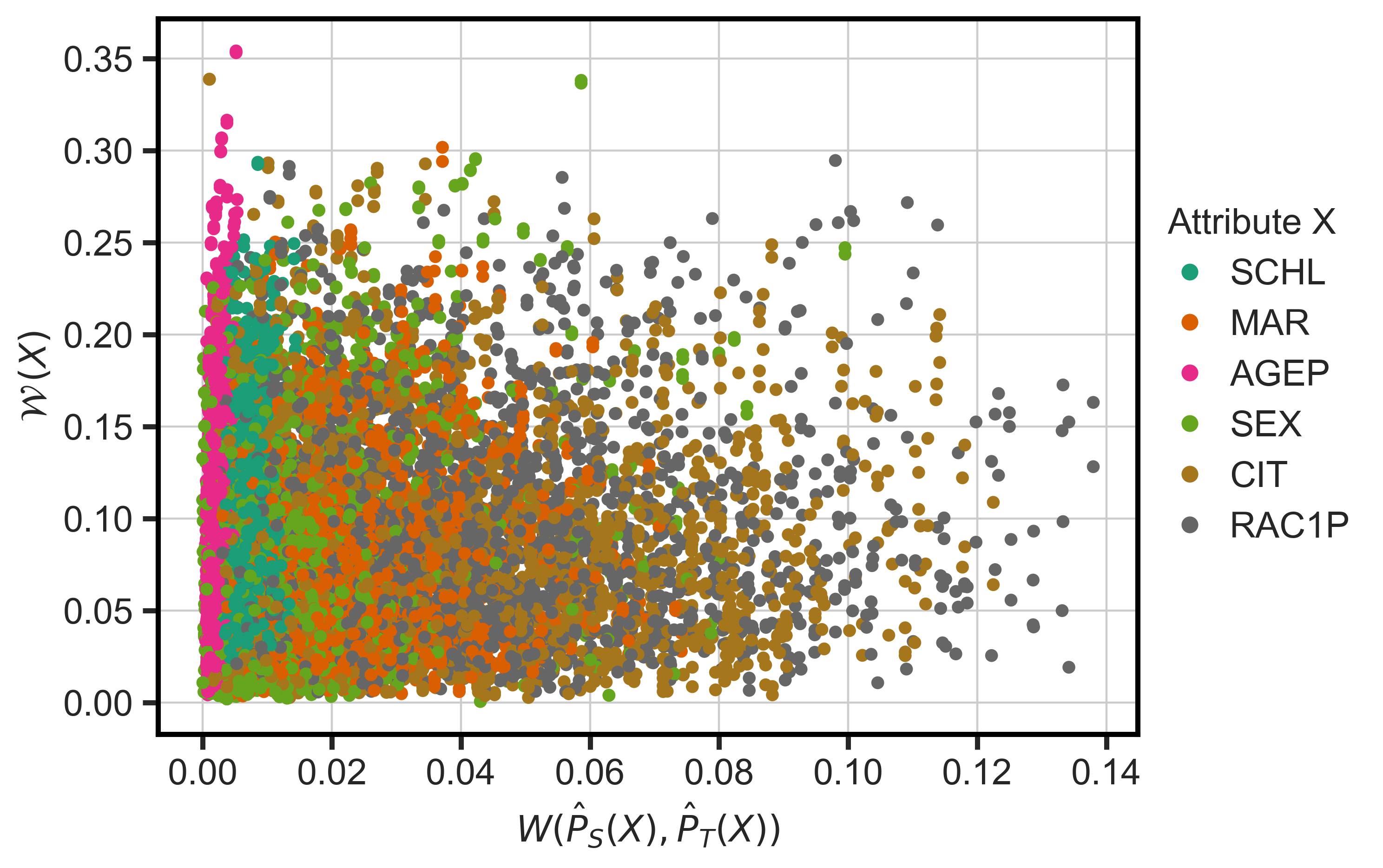}
  \hspace{0.8cm}
  \includegraphics[width=0.34\textwidth]{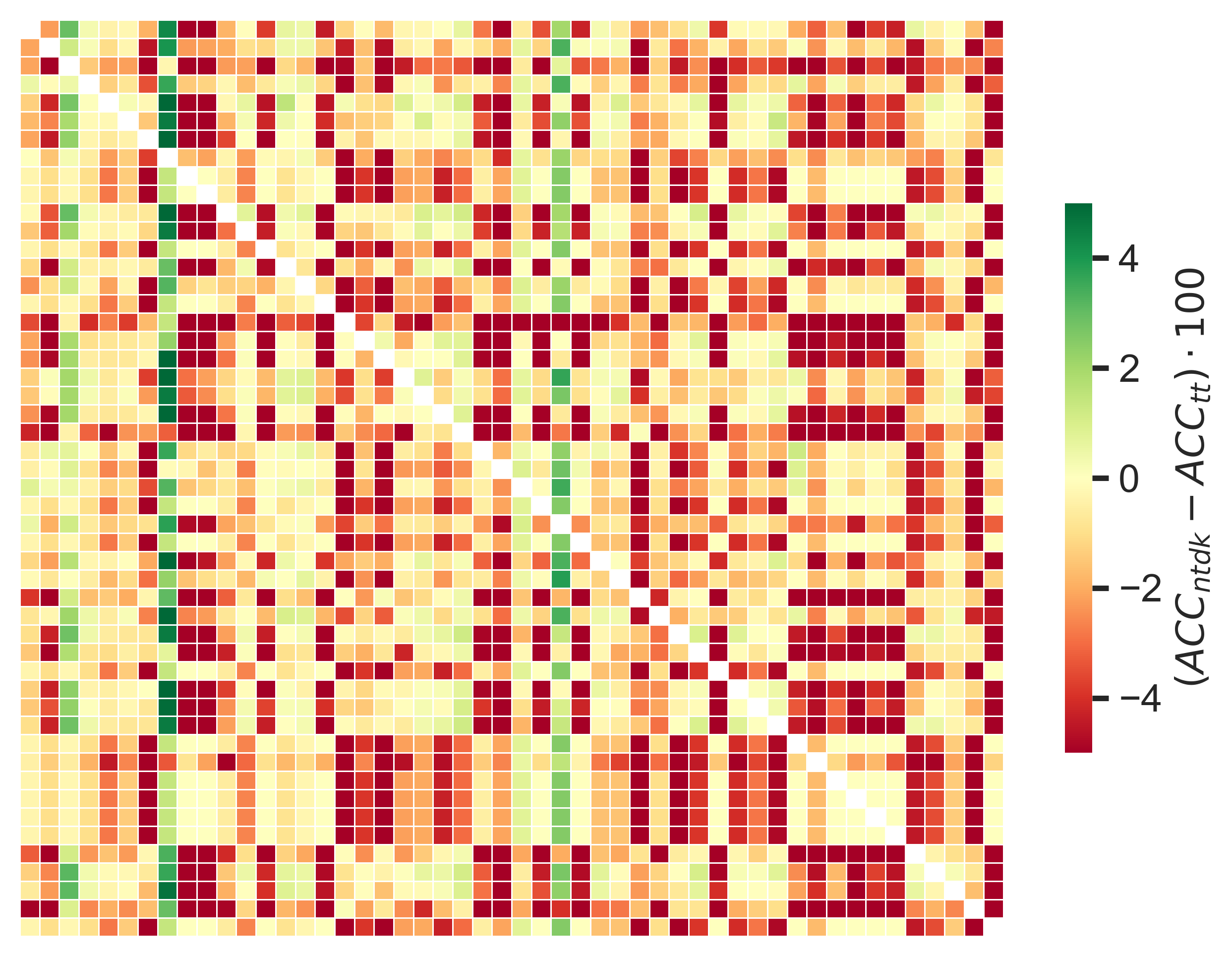}
  \caption{The scatter-plot (left) relates the Wasserstein distances each attribute and source-target US state pair. The x-axis shows the distance of each attribute's marginal distributions between source $\hat{P}_S(X)$ and target domains $\hat{P}_T(X)$, while the y-axis shows the average distance between  conditional $\hat{P}_S(Y|X)$ and $\hat{P}_T(Y|X)$, as in \eqref{eq:wpy}.
  The heat-map (right) shows the difference in accuracy between the cases no target domain knowledge $ACC_{ntdk}$ and target-to-target baseline $ACC_{tt}$ for each source-target US state pair. Both figures show a lack of an overall pattern across all states in \textit{ACSPublicCoverage}. The dataset does not in general satisfy the covariate shift assumption (left).}
  \label{fig:srtg}
\end{figure*}
%

\subsection{Results on Accuracy}
\label{sec:Experiments.CovariateShift}

\label{sec:AccuracyResults}

We now address RQ1 (Section~\ref{sec:ProblemSetting}). The scatter-plot Fig. \ref{fig:srtg} (left) relates the Wasserstein distances for each attribute and source-target pair. On the x-axis, there is the distance between the marginal attribute distributions, i.e., $W(\hat{P}_S(X), \hat{P}_T(X))$. On the y-axis, there is the average distance between conditional $\hat{P}_S(Y|X)$ and $\hat{P}_T(Y|X)$, i.e., $\mathcal{W}(X)$ from \eqref{eq:wpy}.
%
The distances between the marginal attribute distributions are rather small, with the exception of CIT and RAC1P. The distances between class conditional distributions are instead much larger, for all attributes. The plot shows that the \textit{ACSPublicCoverage} dataset does not in general satisfy the covariate shift assumption (at least  when conditioning on a single attribute), but rather the opposite: close attribute distributions and distant conditional class distributions. This fact will help us in exploring how much our approach relies on the covariate shift assumption. Below we report accuracy at varying levels of target domain knowledge (issue P1 Section~\ref{sec:ProblemSetting}), as defined in Section~\ref{sec:DADT:knowledge}.

\textit{\textbf{Case 1: no target domain knowledge (ntdk) vs target-to-target baseline (tt).}} 
Let us consider the scenario of no target domain knowledge, i.e., training a decision tree on the source training data and testing it on the target test data. We compare the decision tree accuracy in this scenario (let us call $ACC_{ntdk}$) to the accuracy of training a decision tree on the target training data and testing on the target test data ($ACC_{tt}$), a.k.a., the target-to-target baseline. Recall that accuracy estimates on a test set (of the target domain) the probability that the classifier prediction $\hat{Y}$ is correct w.r.t. the ground truth $Y$:
\[ ACC = P_T(\hat{Y}=Y) \]

The heat-map plot Fig. \ref{fig:srtg} (right) shows for each source-target pair of states the difference in accuracy $(ACC_{ntdk}-ACC_{tt}) \cdot 100$ between the no target domain knowledge scenario and the
target-to-target baseline. In most of the cases the difference is negative, meaning that there is an accuracy loss in the no target domain knowledge scenario. 

\begin{figure*}[t]
  \centering  
  \includegraphics[width=0.41\textwidth]{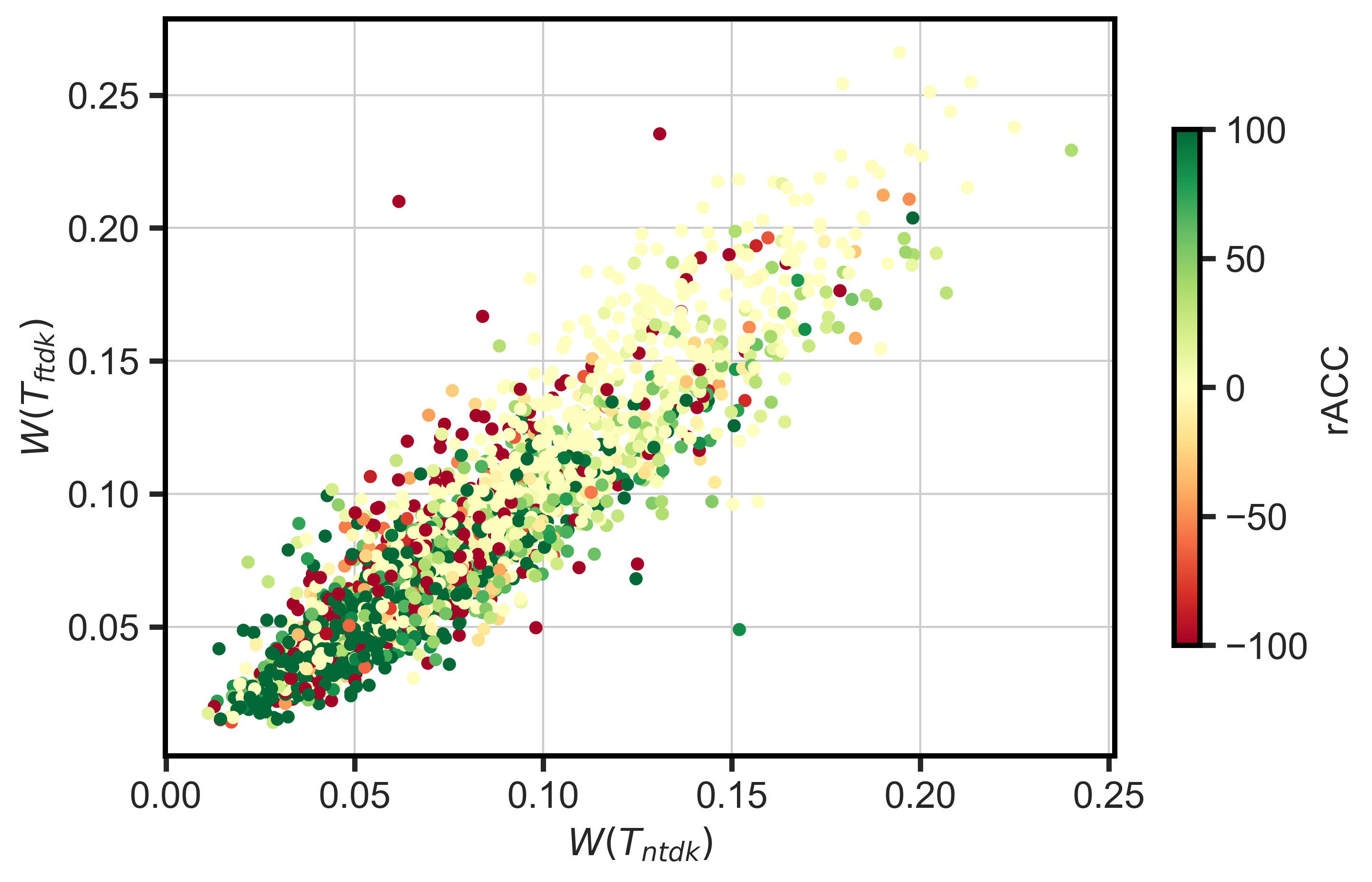}
  \hspace{0.8cm}  \includegraphics[width=0.35\textwidth]{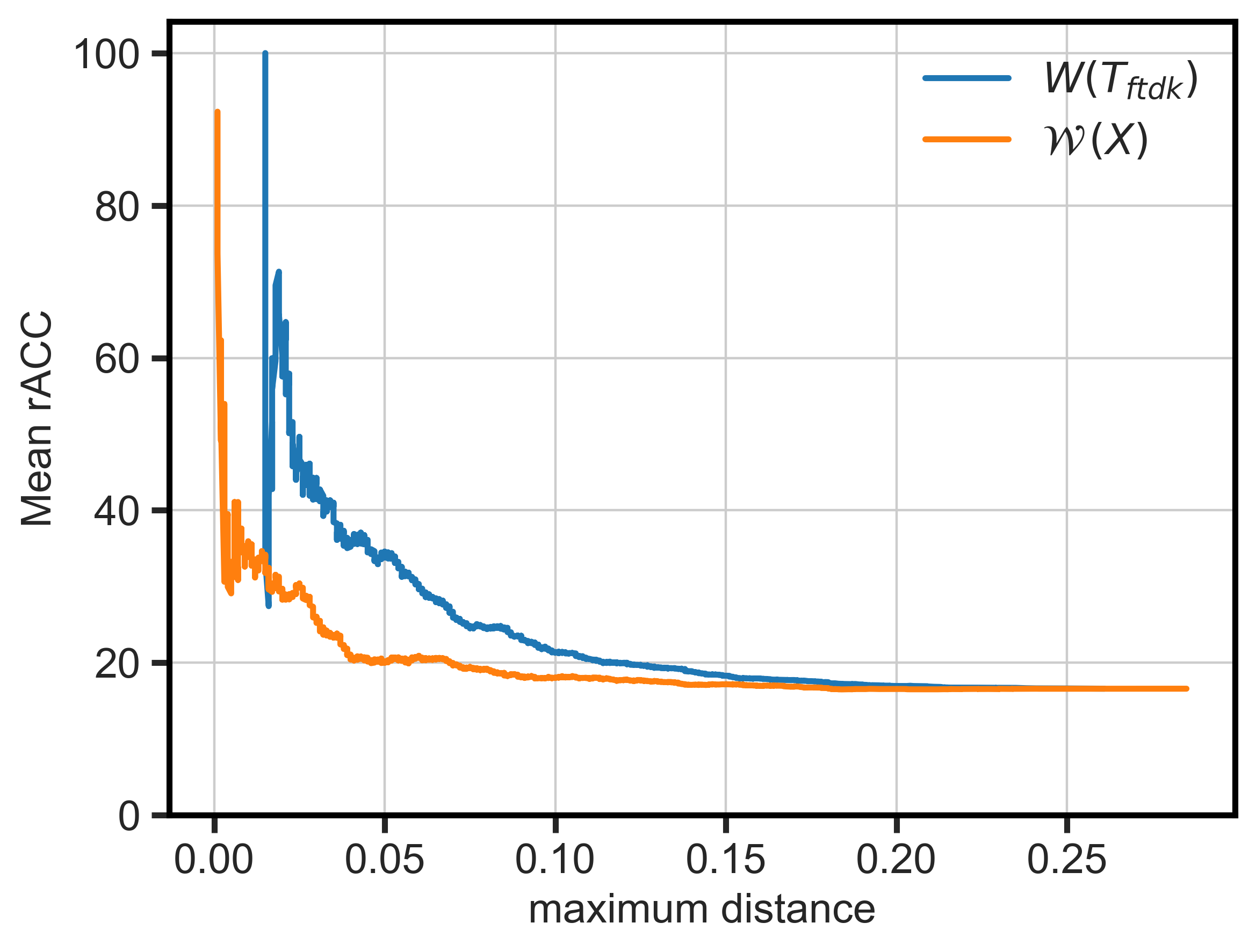}
  \caption{The scatter-plot (left) shows the relative gain in accuracy $rACC$, with a greener dot indicating a greater gain derived from the full target domain knowledge (\textit{ftdk}) relative to the no target domain knowledge (\textit{ntdk}). The x- and y-axis, respectively, shows the covariate shift measured by the Wasserstein distance between the source-target domain pairs used for a decision tree grown in the \textit{ntdk}, $W(T_{ntdk})$, and in the \textit{ftdk}, $W(T_{ftdk})$, scenarios. It shows that a greater gain in accuracy from access to the full target domain knowledge is achieved when the covariate shift assumption is (strictly) met. 
  Similarly the plot (right) shows how model performance (mean $rACC$) deteriorates as the covariate shift assumption is relaxed (shown by a larger Wasserstein distance).}
  \label{fig:res}
\end{figure*}

\textit{\textbf{Case 2: full target domain knowledge (ftdk) vs no target domain knowledge (ntdk).}} The decision tree in this scenario is grown on the source (training) data but probabilities are estimated by full target domain knowledge using (\ref{eq:tdk}), and (\ref{eq:wpyphi}) with $X_w$ minimizing (\ref{eq:wpy}). In the experiments, $\hat{P}_T(X=t|\varphi)$ and 
$\hat{P}_T(X\leq t|\varphi)$ are calculated from the target training data, for each $X$, $t$, and $\varphi$. 

Let us compare the accuracy of the decision tree grown using full target domain knowledge (let us call it $ACC_{ftdk}$) to the one with no target domain knowledge ($ACC_{ntdk}$). In 48\% of the source-target pairs, the accuracy of the full target domain knowledge scenario is better than the one of the no target domain knowledge scenario ($ACC_{ftdk} > ACC_{ntdk}$), and in 26\% of the pairs they are equal ($ACC_{ftdk} = ACC_{ntdk}$). This gross comparison needs to be investigated further. Let us define the relative gain in accuracy as:
\[ rACC = \frac{ACC_{ftdk}-\mathit{min}(ACC_{ntdk},ACC_{tt})}{|ACC_{tt}-ACC_{ntdk}|} \cdot 100 \]
where $ACC_{tt}$ is the accuracy in the target-to-target baseline. The relative gain quantifies how much of the loss in accuracy in the no target domain knowledge scenario has been recovered in the full target domain knowledge scenario. The definition quantifies the recovered loss in accuracy also in the case that $ACC_{ntdk} > ACC_{tt}$, which may occur by chance. Moreover, to prevent outliers due to very small denominators, we cap $rACC$ to the $-100$ and $+100$ boundaries. The mean value of $rACC$ over all source-target pairs is $16.6$, i.e., on average our approach recovers 16.6\% of the loss in accuracy. However, there is a large variability, which we examine further in the next section.

\begin{figure*}[t]
  \centering
  \includegraphics[width=0.35\textwidth]{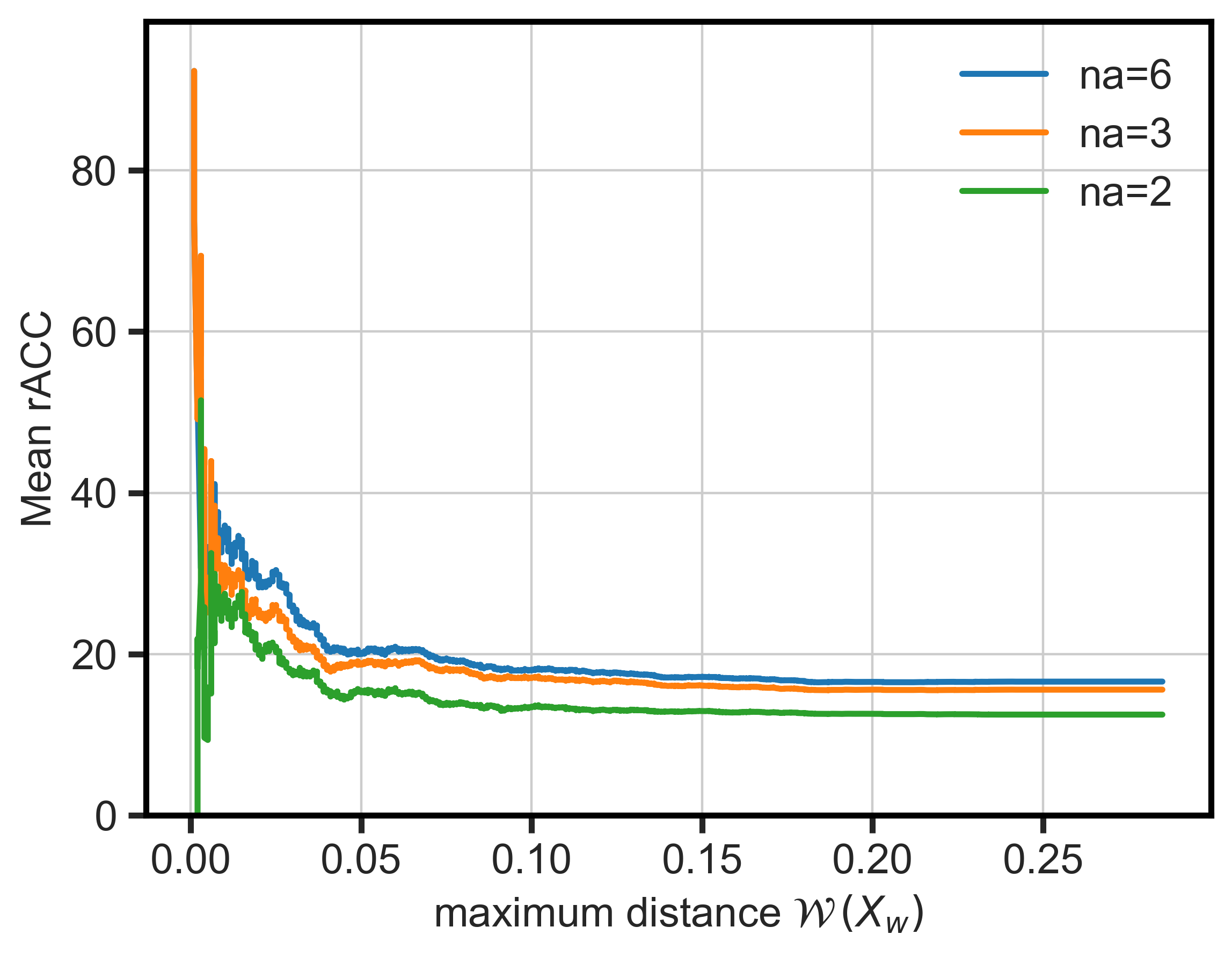}
    \hspace{0.8cm}
    \includegraphics[width=0.35\textwidth]{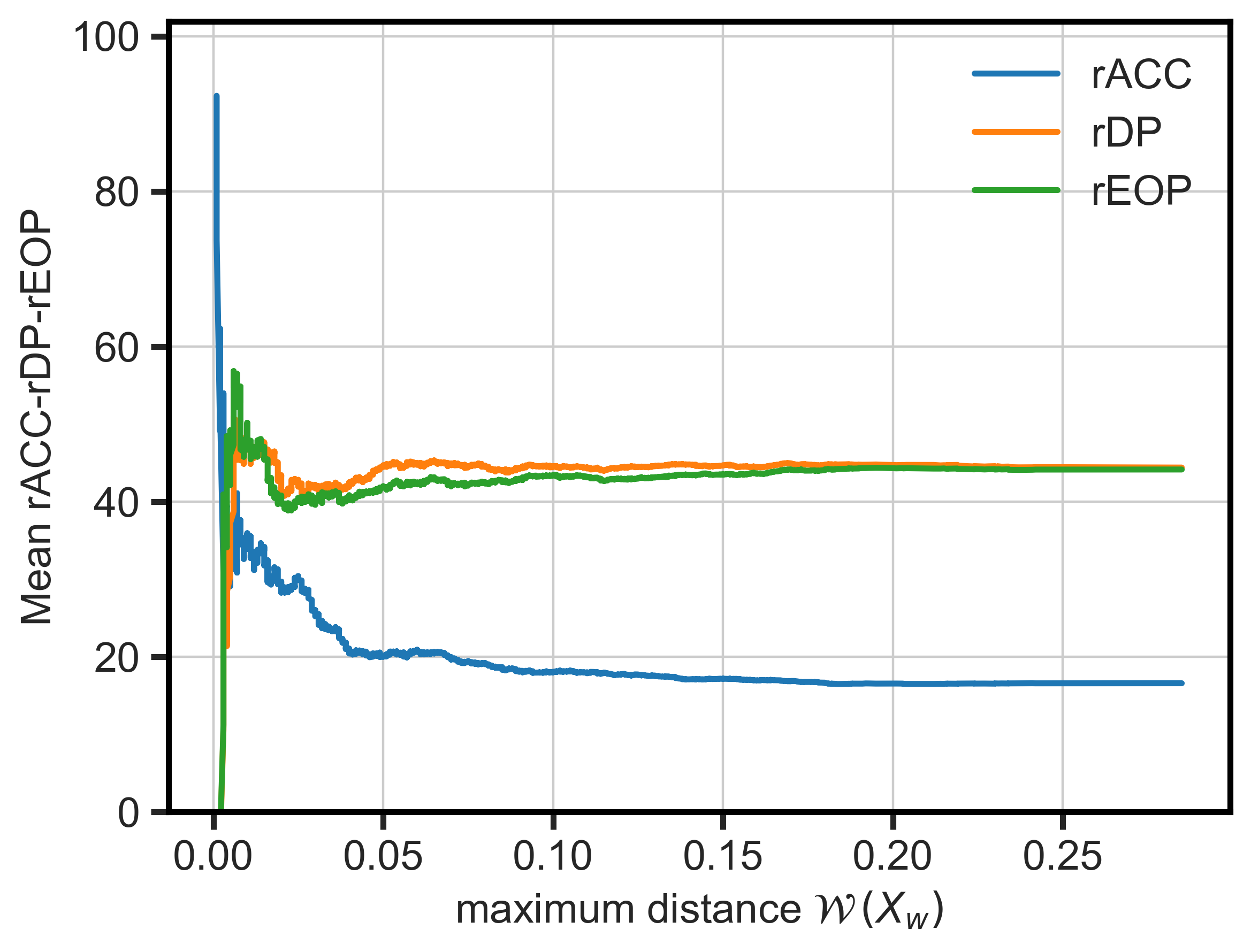}
    \caption{The plot on the left shows results of DADT, across all state pairs, with partial target domain knowledge; we show the mean $rACC$ for pairs with 
    bounded $\mathcal{W}(X_w)$
    for the cases of having knowledge of $na=6$ attributes (i.e., \textit{ftdk}), $na=3$, and $na=2$. The plot on the right shows the change in relative demographic parity, relative equalized odds and accuracy over bounded $\mathcal{W}(X_w)$ in the full target domain knowledge scenario.
    }
  \label{fig:partial}
\end{figure*}

\textit{\textbf{Case 3: partial target domain knowledge.}}
We reason on partial target domain knowledge under the assumption that we only know an estimate of the distribution of some subsets of $\mathbf{X}$'s but not of the full joint  probability distribution $P_T(\mathbf{X})$. We experiment assuming to know $\hat{P}_T(\mathbf{X}')$ for $\mathbf{X}' \subseteq \mathbf{X}$, only if $|\mathbf{X}'| \leq 2$ (resp., $|\mathbf{X}'| \leq 3$). Equivalently, we assume to know $\hat{P}(X=x|\varphi')$ only if $\varphi'$ contains at most one (resp., two) variables. Formulas \eqref{eq:aff}--\eqref{eq:aff2} mix such a form of target domain knowledge with the estimates on the source: for $\hat{P}(X=x|\varphi)$, we compute $\varphi'$ as the subset of split conditions in $\varphi$ regarding at most the first (resp., the first two) attributes in $\varphi$ -- namely, the attributes used in the split condition at the root (resp., at the first two levels) of the decision tree, which are the most critical~ones.
The weight $\alpha$ in \eqref{eq:aff}--\eqref{eq:aff2} is set dynamically as the proportion of attributes in $\varphi$ which are not in $\varphi'$. This value is $0$ when $\varphi$ tests on at most one variable (resp., two variables), and greater than $0$ otherwise. We consider the proportion of attributes and not of the number of split conditions, since continuous attributes may be used in more than one split along a decision tree path.


\textit{\textbf{Covariate Shift and Accuracy.}}
We test whether the difference in model performance is due to the fact that different pairs match or do not match the covariate shift assumption. In order to quantify the covariate shift (issue P2), we define specifically for a decision tree $T$:
\begin{align}
\label{eq:wtree}
    W(T) = \sum_{\varphi\ \mbox{\rm path of a leaf of\ } T} W( \hat{P}(Y|\varphi), \hat{P}_T(Y|\varphi)) \cdot \hat{P}_T(\varphi)
\end{align}
as the average Wasserstein distance between the estimated (through \eqref{eq:wpyphi}) and target domain class distributions at leaves of the decision tree, weighted by the leaf probability in the target domain. Notice that, as $P_T$ is unknown, we estimate the probabilities in the above formula on the test set of the target domain. We write $W(T_{ntdk})$ and $W(T_{ftdk})$, respectively, for denoting the amount of covariate shift for the decision tree grown in the no target domain knowledge and with full target domain knowledge scenarios.
The scatter plot Fig. \ref{fig:res} (left) shows the relative accuracy (in color) at the variation of $W(T_{ntdk})$ and $W(T_{ftdk})$\footnote{$W(T_{ntdk})$ and $W(T_{ftdk})$ appear to be correlated. While they are specific of their respective decision trees, they both depend on the distribution shift between the source and target domain.}. We make the following qualitative observations:
\begin{itemize}
    \item when $W(T_{ftdk})$ is small, say smaller than $0.05$, i.e., when the covariate shift assumption holds, the relative accuracy is high, i.e., using target domain knowledge allows for recovering the accuracy loss;
    \item when $W(T_{ftdk})$ is large, in particular, larger than $W(T_{ntdk})$,    
    then the gain is modest or even negative.
\end{itemize}

Let us consider how to determine quantitatively on which pairs there is a large relative accuracy. Fig.~\ref{fig:res} (right) reports the mean $rACC$ for source-target pairs sorted by two different distances. Ordering by $W(T_{ftdk})$ allows to identify more source-target pairs for which our approach works best than ordering by the average class conditional distance $\mathcal{W}(X_w)$, where $X_w$ is
%
from (\ref{eq:wpy}). However:
\begin{itemize}
    \item $W(T_{ftdk})$ requires target domain knowledge on $P_T(Y|\varphi)$ for each 
    leaf in $T_{ftdk}$, which is impractical to obtain.
    \item $\mathcal{W}(X_w)$ 
    is easier to calculate/estimate, as it regards only the conditional 
    distribution $P_T(Y|X)$. The exact knowledge of which attribute is $X_w$ is not required, as, by definition of $X_w$, using any other attribute instead of $X_w$ provides an upper bound to $\mathcal{W}(X_w)$. 
\end{itemize}
In summary, Fig. \ref{fig:res} (right) shows that DADT is able to recover a good proportion of loss in accuracy, and it provides a general guidance for selecting under how much the covariate shift assumption can be relaxed. Finally, Fig. \ref{fig:partial} (left) contrasts the $rACC$ metric of the full target domain knowledge scenario to the two cases of  the partial target domain knowledge scenario when we have knowledge of only pairs or triples of variables. There is, naturally, a degradation in the recovery of accuracy loss in latter scenarios, e.g., for a distance of up to $0.03$, we have the mean $rACC$ equal to $25.3\%$ for full target domain knowledge, to $21.6\%$ when using triples, and to $17.5\%$ when using pairs of variables\footnote{The extension of $rACC$ to partial target domain knowledge is immediate by replacing $ACC_{ftdk}$ in its definition with the accuracy $ACC_{ptdk}$ of the decision tree grown by using partial target domain knowledge.}. 
Even with partial target domain knowledge in the form of cross-tables, we can achieve a moderate recovery of the loss in accuracy.

\begin{figure*}[t]
  \centering
  \begin{tabular}{cccc}
  \hspace{-1ex}
  \includegraphics[width=0.231\linewidth]
   {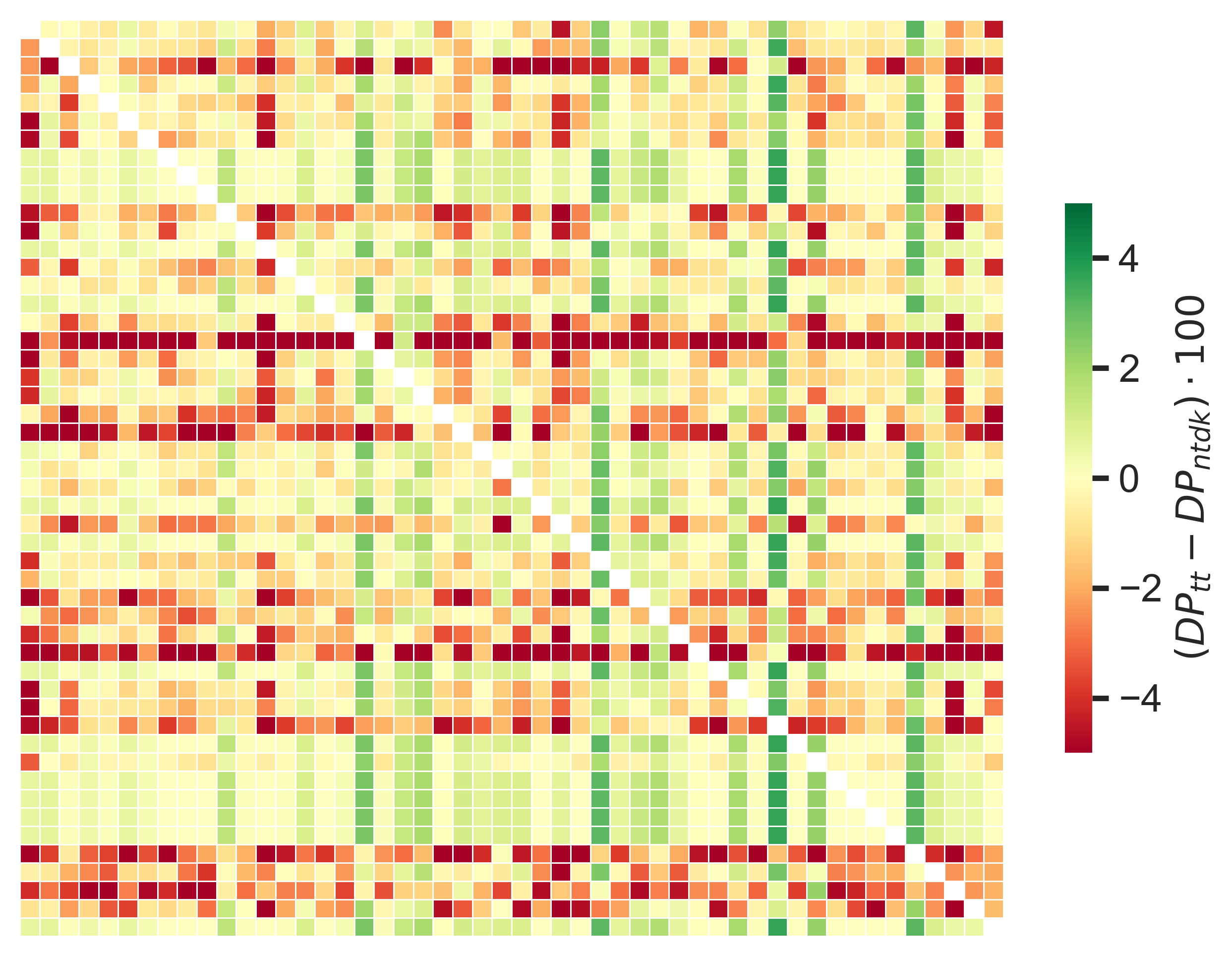}&
   \includegraphics[width=0.233\linewidth]
  {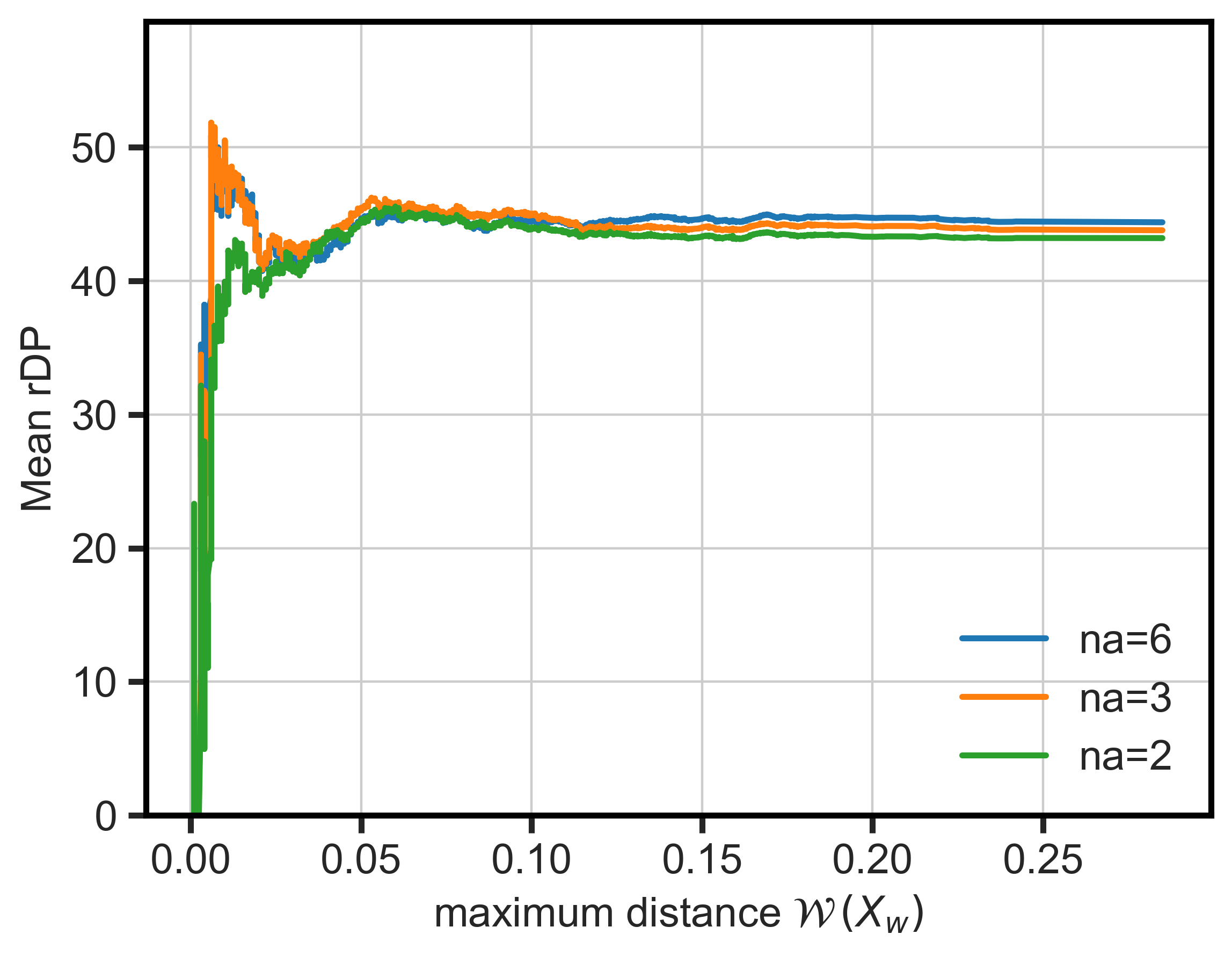}&
  \includegraphics[width=0.231\linewidth]
   {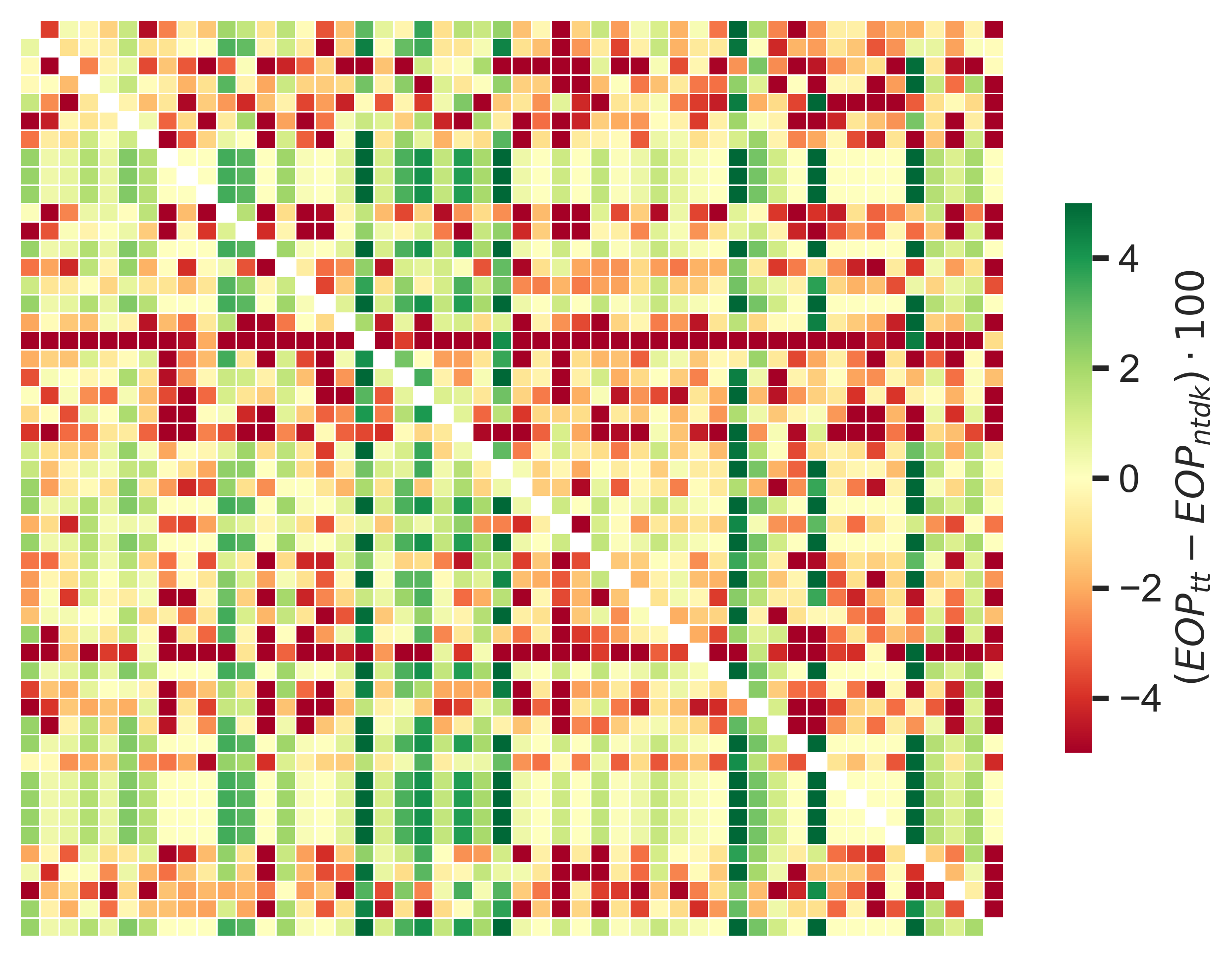}&
   \includegraphics[width=0.233\linewidth]
  {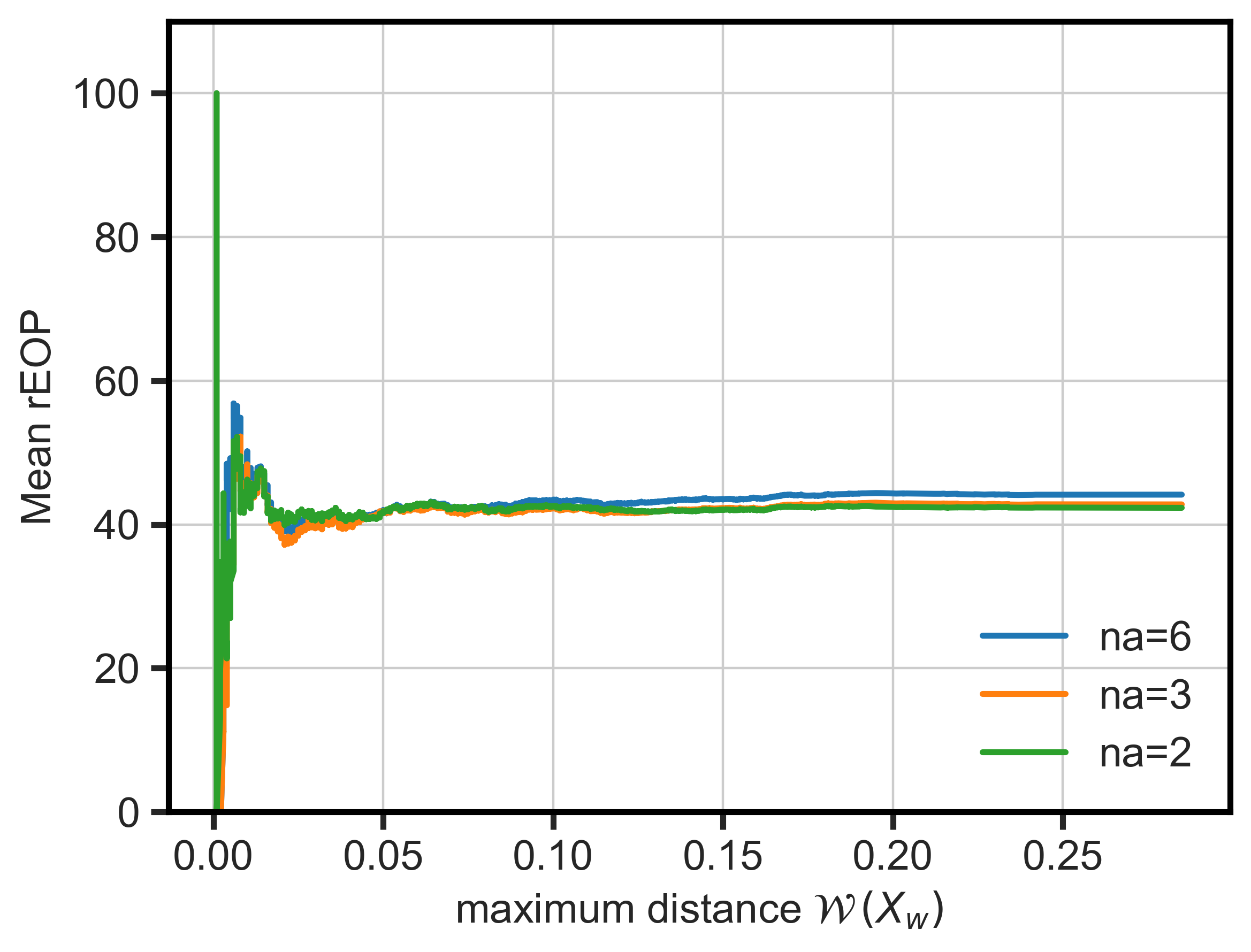}\\
  (a) & (b) & (c) & (d)\\
    \end{tabular}
  \caption{
  The heatmap (a) shows the difference in DP  between the cases target-to-target baseline $DP_{tt}$ and no target domain knowledge $DP_{ntdk}$ for each source-target US state pair. Similarly (c) for EOP.
  The plots (b) and (d) show the mean $rDP$  and $rEOP$ respectively, for the cases of having knowledge of $na=6$ attributes (i.e., \textit{ftdk}), $na=3$, and $na=2$, over  bounded $\mathcal{W}(X_w)$. 
  }
  \label{fig:resdp}
\end{figure*}

\subsection{Results on Fairness}
\label{sec:fairnessresults}

We now address experiments on RQ2 (Section~\ref{sec:ProblemSetting}). Other quality metrics beyond accuracy can degrade in presence of covariate shift. There is also a risk that certain demographic groups are more impacted by drops in accuracy than others. This can occur even if overall minimal accuracy drop is seen. In order to test the impact of DADT on specific groups, and answer RQ2, we utilize two \emph{fairness metrics} commonly used in fair machine learning literature, \emph{demographic parity} and \emph{equal opportunity}. We consider here the fairness metrics in reference to the protected attribute SEX. We study how DADT compares to the standard decision tree under the same post-processing fairness step. We hypothesize that under a DA scenario, said step is more impactful under a DADT classifier as it can account for the target domain information during training.
\textit{Demographic parity} (DP) quantifies the disparity between predicted positive rate for men and women: 
\[ DP = |P(\hat{Y}=1|\mbox{SEX=women})-P(\hat{Y}=1|\mbox{SEX=men})| \]

The lower the $DP$ the better is the fairness performance. We consider this metric in the context of our women's football shoes example, where one measure of whether a model is addressing our identified feedback loop is whether the positive rate for the question of ``will buy football shoes'' moves towards parity for men and women.
\textit{Equal opportunity} (EOP) quantifies the disparity between true positive rate for men and women:
\[EOP = |P(\hat{Y}={y}|\mbox{SEX=women}, {Y}={1})-P(\hat{Y}=1|\mbox{SEX=men}, {Y}={1})| \]
We consider this metric in the context of the example of the public administrator who is identifying people who do not receive benefits to which they are entitled to. Here, our concern is that the model is equally performant for all groups as prescribed by SEX.


Fairness-aware classifiers control for these metrics. We use here a classifier-agnostic post-processing method, described in Section~\ref{sect:expsetup} that specializes the decision threshold for each protected group \cite{DBLP:conf/nips/HardtPNS16}. The correction is applied after the decision tree is trained. Fig. \ref{fig:resdp} (a) confirms a degradation of the DP metric from the target-to-target scenario to the no target domain knowledge scenario. Fig. \ref{fig:resdp} (c) shows a less marked degradation for the EOP metric.

We mimic the reasoning done for the accuracy metric in Section \ref{sec:AccuracyResults}, and introduce the relative gain in demographic parity ($rDP$) and the relative gain in equal opportunity ($rEOP$):
\begin{align*}
    rDP =& \frac{\mathit{max}(DP_{ntdk}, DP_{tt})-DP_{ftdk}}{|DP_{tt}-DP_{ntdk}|} \cdot 100 \\
    rEOP =& \frac{\mathit{max}(EOP_{ntdk}, EOP_{tt})-EOP_{ftdk}}{|EOP_{tt}-EOP_{ntdk}|} \cdot 100
\end{align*}
Note that since DP and EOP improve when they become smaller, the definitions of relative gain are symmetric if compared to the one of $rACC$.

%
%

Fig. \ref{fig:partial} (right) substantiates also for $rDP$ and $rEOP$ the conclusions for $rACC$ mentioned in Section~\ref{sec:fairnessresults}. The distance $\mathcal{W}(X_w)$
turns out to provide a guidance on when DADT works the best. For DP and EOP, however, for large values of such a distance, we do not observe a degradation as in the case of ACC. In other words, when the assumption of covariate shift is strictly met, DADT works the best, but when it is not, the recovery of the DP and EOP does not degrade.

Finally, Fig. \ref{fig:resdp} (b)  confirms the degradation of the DADT performances in the case of partial target domain knowledge. E.g., for a distance of $0.03$ we have the mean $rDP$ equal to $41.8\%$ for full target domain knowledge, $42.2\%$ when using triples, and $40.8\%$ when using pairs of variables. This is much less marked for $rEOP$, for which DADT performs very well also with knowledge of pairs of variables, as shown in Fig. \ref{fig:resdp} (d).

\section{Related Work}
\label{sec:RelatedWork}
 
Domain adaptation (DA) studies how to achieve a robust model when the training (source domain) and test (target domain) data do not follow the same distribution \cite{Redko2020_DASurvey}. Here, we focused on the covariate shift type, which occurs when the attribute space $\mathbf{X}$ is distributed differently across domains \cite{Quinonero2009,DBLP:journals/pr/Moreno-TorresRACH12, DBLP:conf/icml/ZhangSMW13}. To the best of our knowledge, DADT is the first framework to address domain DA as an in-processing problem specific to decision tree classifiers.

Previous work on adjusting entropy estimation has been conducted largely outside of machine learning, as well as in the context of information gain (IG) in decision trees. Here, too, DADT is the first work to look at entropy estimation under DA. \citet{guiasuweighted1971} proposes a general form for a weighted entropy equation for adjusting the likelihood of the information being estimated. Other works study the estimation properties behind using frequency counts for estimating the entropy \cite{Schurmann_2004, DBLP:conf/icml/Nowozin12, DBLP:conf/nips/NemenmanSB01, DBLP:journals/jmlr/ArcherPP14}. In relation to decision trees, \cite{DBLP:journals/eswa/SingerAB20} proposes a weighted IG based on the the risk of the portfolio of financial products that the decision tree is trying to predict. Similarly, \cite{DBLP:conf/ijcai/ZhangN19, DBLP:conf/dis/ZhangB20} re-weight IG with the fairness metric of statistical parity. \citet{DBLP:conf/csemws/VieiraA14} adjust the IG calculation with a gain ratio calculation for the purpose of correcting a bias of against attributes that represent higher levels of abstraction in an ontology.

Recent work has started to examine the relationship between DA and fairness. \citet{DBLP:conf/nips/MukherjeePYS22} show that domain adaptation techniques can enforce individual fairness notions. \citet{maityDoesEnforcingFairness} show that enforcing risk-based fairness minimization notions can have an ambiguous effect under covariate shift for the target population, arguing that practitioners should check on a per-context basis whether fairness is improved or harmed. This is line with the findings of \cite{dingRetiringAdultNew2021} who test both standard and fairness adjusted gradient boosting machines across numerous shifted domains and find that both accuracy and fairness metric measures are highly variable across target domains. These works call for further work to understand the impact of domain drifts and shifts.

%
%

%
\section{Conclusion}
\label{sec:Closing}

In answer to RQ1 and RQ2 (Section~\ref{sec:ProblemSetting}), we see that domain-adaptive decision trees (DADT) result in both increased accuracy and better performance on fairness metrics over our baseline standard decision tree trained in $D_S$ and tested in $D_T$. Looking more closely at our experimental results, we see that improvements are best when the covariate shift assumption holds in at least a relaxed form (P2). We also see this increase when we only have partial domain knowledge (P1), though a greater amount of domain knowledge, as we define it, results in greater improvements in those metrics. Interestingly, our post-processing fairness intervention does not have worse performance over a standard decision tree even when the covariate shift assumption does not hold. 

Back to the example inspired by the experimental setting in Section~\ref{sec:Experiments}, we have demonstrated that DADTs are an effective method for using existing information about a target state. We can also think back to our retail example in Section~\ref{sec:Introduction}, wherein we identified a potential feedback loop leading to a lack of stock in women's football shoes. We propose that DADTs are a method for intervening on this feedback loop; if the store identified a pool of potential customers (such as the population living near the store), which had a higher rate of women than their existing customer base, DADT provides an accessible, interpretable, and performative classification model which can incorporate this additional information. In future work, different definitions of outside information should be explored as the outside information may not have the same structure as the source and target datasets.

While we see that the benefits are clear, we want to be explicit about the limitations of our method. Firstly, we show that this is most effective when the covariate shift assumption holds. We consider a strength of our work that we specify and test this assumption and encourage future work on domain adaptation methods to similarly specify the conditions under which a method is suitable to be used. Secondly, we emphatically acknowledge that DADTs are not intended as a replacement for collecting updated and improved datasets. However, this is a low cost improvement that can be made over blindly applying to a new or changing context. Additionally, there are cases in which labelled data simply does not exist yet. Finally, DADTs are not a complete solution for achieving or ensuring fair algorithmic decision making; rather they are an easy to use method for improving accuracy, and fairness metric performance in the commonly occurring case of distribution shift between source and target data. 

%
%

\begin{acks}
    This work has received funding from the European Union’s Horizon 2020 research and innovation program under Marie Sklodowska-Curie Actions (grant agreement number 860630) for the project "NoBIAS - Artificial Intelligence without Bias". This work reflects only the authors' views and the European Research Executive Agency (REA) is not responsible for any use that may be made of the information it contains.
\end{acks}

\bibliographystyle{ACM-Reference-Format}
\bibliography{references_condensed}

\newpage

\appendix
\section{Supplementary Material}

\subsection{Distance between Probability Distributions}
\label{sec:sec:ProblemSetting.DistBtwProbs}

We resort to the \textit{Wasserstein distance} $W$ between two probability distributions to quantify the amount of covariate shift and the robustness of target domain knowledge. In the former case, we quantify the distance between $P_S(Y|\mathbf{X})$ and $P_T(Y|\mathbf{X})$. In the latter case, the distance between $P_S(X|\varphi)$ and $P_T(X|\varphi)$. 
We define $W$ between $P_S$ and $P_T$ as:
\begin{equation*}
\label{eq:WDistance}
    W(P_S, P_T) = \int_{-\infty}^{+\infty} | \mathcal{P}_S - \mathcal{P}_T |
\end{equation*}
where $\mathcal{P}_S$ and $\mathcal{P}_T$ are the cumulative distribution functions (CDFs) of $P_S$ and $P_T$.\footnote{See \url{https://docs.scipy.org/doc/scipy/reference/generated/scipy.stats.wasserstein_distance.html} for implementation details.} We can estimate $\mathcal{P}_S$ and $\mathcal{P}_T$ from the data using 
(6).
The smaller $W$ is, the closer are the two distributions, indicating similar informational content.

\subsection{Additional Theoretical Discussion}
\label{Appendix:SuppMaterial}

Recall the equality 
(10),
which is central to covariate shift. Under a decision tree learning setting, it does not necessarily imply $P_T(Y=y|\varphi) = P_S(Y=y|\varphi)$ for a current path $\varphi$. Consider the example below.

\begin{example}
\label{example:EstYUnderCovShift}
Let $\mathbf{X} = X_1, X_2$ and $Y$ be binary variables, and $\varphi$ be $X_1=0$. Since $P(X_1, X_2, Y) = P(Y|X_1, X_2) \cdot P(X_1, X_2)$, the full distribution can be specified by stating $P(Y|X_1, X_2)$ and $P(X_1, X_2)$. Let us consider any distribution such that:
\[P_S(X_1, X_2) = P_S(X_1) \cdot P_S(X_2) \quad P_T(X_1=X_2)=1 \quad Y = I_{X_1=X_2} \]
i.e., $X_1$ and $X_2$ are independent in the source domain, while they are almost surely equal in the target domain. Notice that $Y = I_{X_1=X_2}$ readily implies that $P_S(Y|X_1, X_2) = P_T(Y|X_1, X_2)$, i.e., the covariate shift condition 
(10)
holds. Using the multiplication rule of probabilities, we calculate:
\begin{align*}
    P_S(Y|\varphi) = P_S(Y|X_1=0) &= \\
    P_S(Y|X_1=0, X_2=0) &\cdot P_S(X_2=0|X_1=0) \; + \\
        P_S(Y|X_1=0,& X_2=1) \cdot P_S(X_2=1|X_1=0) = \\
    P_S(Y|X_1=0, X_2=0) &\cdot P_S(X_2=0) \; + \\
        P_S(Y|X_1=0,& X_2=1) \cdot P_S(X_2=1)
\end{align*}
where we exploited the independence of $X_1$ and $X_2$ in the source domain, and
\begin{align*}
    P_T(Y|\varphi) = P_T(Y|X_1=0) &= \\
    P_T(Y|X_1=0, X_2=0) &\cdot P_T(X_2=0|X_1=0) \; + \\
        P_T(Y|X_1=0,& X_2=1) \cdot P_T(X_2=1|X_1=0) = \\
    P_T(Y|X_1=0, X_2=0) &
\end{align*}
were we exploited the equality of $X_1$ and $X_2$ in the target domain.
$P_S(Y|\varphi)$ and $P_T(Y|\varphi)$ are readily different when setting $X_1, X_2 \sim Ber(0.5)$ because $P_S(Y=1|\varphi) = 1 \cdot 0.5 + 0 \cdot 0.5 \neq 1 = P_T(Y=1|\varphi)$. 
\end{example}

%
%

\end{document}